# An Unsupervised Feature Learning Approach to Improve Automatic Incident Detection

Jimmy SJ. Ren, Wei Wang, Jiawei Wang, and Stephen Liao

*Abstract*— sophisticated automatic incident detection (AID) technology plays a key role in contemporary transportation systems. Though many papers were devoted to study incident classification algorithms, few study investigated how to enhance feature representation of incidents to improve AID performance. In this paper, we propose to use an unsupervised feature learning algorithm to generate higher level features to represent incidents. We used real incident data in the experiments and found that effective feature mapping function can be learnt from the data crosses the test sites. With the enhanced features, detection rate (DR), false alarm rate (FAR) and mean time to detect (MTTD) are significantly improved in all of the three representative cases. This approach also provides an alternative way to reduce the amount of labeled data, which is expensive to obtain, required in training better incident classifiers since the feature learning is unsupervised.

## I. INTRODUCTION

Many countries around the world suffer from increasing traffic congestions in their freeway networks. Recent report showed that traffic incident is one of the major contributors to traffic congestions. According to [1], 25% of the traffic congestions in the U.S. were caused by traffic incident. This number is 12% for Paris metropolitan region and 33% for German motorways. Delays in incident detection may not only deteriorate the traffic, it may also make the post-incident management such as emergency services and security services more difficult. Therefore, an automatic, high performance and reliable incident detection system is vital for both traffic reasons and safety reasons. Automatic incident detection (AID) is an area which has been investigated for several decades. It is well recognized that inductive loop is one of the most widely used technologies in practice to generate AID data [2]. Moreover, this technology is not only well understood but also provides an accurate and economical method for incident detection [1]. In this study, we focus on new AID method which is based on inductive loop data.

Many studies have been devoted to algorithms improving AID by using inductive loop data. Among many kinds of algorithms studied, machine learning algorithms were most investigated in recent years. A number of variations of artificial neural network (ANN) [3][4][5][6], support vector machine (SVM) [7][8][9][10] and other algorithms were proposed and tested in the AID context. Significant improvements were obtained in the field by using machine learning methods. There are two common characteristics among most of such studies. Firstly, for the feature representation which represents a detection moment at a particular area, only very limited amount of data was used. Secondly, they all use raw data generated by inductive loops in both upstream detectors and downstream detectors as feature representations for incident classification. This naturally introduced two potential issues. One is even if more data is available from detectors; it's not clear whether by taking advantage of them shall benefit the incident detection performance. The other one is they only used raw feature representation and overlooked the higher level feature representation which may potentially better represent the incidents. Since much recent work in machine learning focused on unsupervised feature learning which learns high quality higher level feature representation from unlabeled data and such methods yielded promising results in the tasks such as classification [14][17][20], we believe such approach can fill this research gap in AID area. In this study we address the first issue by looking at how different choices of raw features influence the AID performance. Then we use unsupervised feature learning to learn higher level feature representation to improve incident detection performance.

We carried out the experiment against the real traffic incident data and our result shows that there is a trade-off between false alarm rate (FAR) and mean time to detect (MTTD) when choosing different number of intervals (cycles) included in the raw feature. We also show that by adding higher level features learnt by the unsupervised feature learning, AID performance is significantly improved compare to all of the three representative raw feature choices. We also illustrate that this method is able to learn excellent feature mapping function from unlabeled data of a different test site. This opens a door to improve AID performance even if data from the current test site is not sufficient.

The organization of this paper is the following. Section 2 reviews the literature in both AID algorithms and unsupervised feature learning; then identifies the research gap. Section 3 introduces the unsupervised feature learning algorithm and research methodology. Section 4 shows and analyzes the experiment results. Section 5 summarizes the contributions and the limitations; then proposes future research directions.

## II. LITERATURE REVIEW

### A. Machine Learning in Automatic Incident Detection (AID)

Many AID studies adopted machine learning methods since 1990s. Artificial neural network (ANN) was widely applied in the 90s. Many variations of ANN can be found in the literature. Cheu and Ritchie [3] compared three types of ANNs namely multi-layer feed-forward (MLF) neural

Jimmy SJ. Ren is with Department of Information Systems, City University of Hong Kong (e-mail: sjren2@student.cityu.edu.hk).
Wei Wang is with Department of Information Systems, City University of Hong Kong (e-mail: wewang8@student.cityu.edu.hk).
Jiawei Wang is with CityU-USTC Joint Research Center (e-mail: wangjw2@mail.ustc.edu.cn).
Stephen Liao is a professor of Information Systems, City University of Hong Kong (e-mail: issliao@cityu.edu.hk).

network, self-organizing feature map (SOFM) and adaptive resonance theory 2 (ART2). They found that MLF neural network performed best in terms of incident detection performance. It was also reported in [4] that constructive probabilistic neural network (CPNN) is able to achieve good AID performance with a much smaller network size, and it enjoys better application potential than basic probabilistic neural network (BPNN). In [5], the authors used genetic algorithm to optimize the parameter search for probabilistic neural network (PNN) and the result showed that this approach achieved better DR and FAR. A modified form of the Bayesian-based PNN was proposed in [6] and the study result indicated that this method enhances the universality and transferability of the detection algorithm.

More recent studies focused on the application of support vector machine (SVM) in improving AID performance. In [7] the authors found that SVM generated better performance index (PI), a systematic measure of AID performance, than ANN. SVMs with polynomial kernel and radial basis function (RBF) kernel were compared in [8] and it was found that SVMs generate lower misclassification rate (MCR), higher DR and lower FAR compare to MLF neural network and PNN. By carefully selecting algorithm parameters, [9] shows SVM is able to outperform traditional AID algorithms. In [10], the authors used different strategies to combine SVMs to avoid the burden of choosing kernel functions and tuning the parameters. Result showed that this approach performs better than single SVM based AID algorithm in many aspects. Though other machine learning methods were also used in some studies, we found most of the AID research in the past decade applied ANNs and SVMs and made significant progress in terms of AID performance. It is also clear that among recent AID studies using machine learning methods, most of them used raw features which include direct values of the detection station average data accumulated over 30 seconds in the classification. Up to five past intervals from upstream detectors and three past intervals from downstream detectors were usually adopted to represent one detection moment. By this convention, the resulting dimensionality of the features representing incident/non-incident moments is fairly low.

*B. Recent Advances in Unsupervised Feature Learning*

Learning internal representation of raw features is not a new idea. When back propagation algorithm was proposed to solve MLF neural network, it was explicitly indicated in [11] that the hidden layer is the internal representation of the input features. Principle component analysis (PCA) is another method seeks internal structure of data and is widely used in many unsupervised learning tasks. But none of these internal representations was endowed any semantic interpretation. Breakthrough was made in [12]. The authors showed that by maximizing the sparseness in the coding strategy, unsupervised learning algorithm is able to learn a feature mapping function which generates a complete set of localized, oriented receptive fields from natural images. These receptive fields are not only similar to those found in the primary visual cortex of mammals; they are also more effective feature representation for later tasks such as classification due to their higher degree of statistical independence. Since the process of learning the feature mapping function is purely unsupervised, such procedure is called unsupervised feature learning or unsupervised pre-training in general.

Many other unsupervised feature learning methods were developed thereafter to learn multiple layers of higher level feature representation rather than a single layer. By stacking layers of features, deep architecture can be formed. Off-the-shelf unsupervised feature learning algorithms include sparse-coding [13], Restricted Boltzmann Machine (RBM) [14], sparse auto-encoder [15], denoising auto-encoder [16], K-means [17], etc. Such methods are also collectively called deep learning methods. Many recent studies reported that not only in image classification, state-of-the-art performance in other tasks like audio classification and text classification can also be achieved by using features generated by applying unsupervised feature learning in audio and text data [18][19].

In sum, unsupervised feature learning algorithms are able to learn feature mapping functions which map raw features (such as raw image pixels and audio frequencies) to higher level features. The higher level features are to be the better feature representation for later tasks such as classification. Though most of the studies focused on human perceptible data (e.g. image, audio and text), unsupervised feature learning is a general framework which can be applied to any data [20].

*C. Feature Engineering in AID and Research Gap*

To our knowledge, very few studies in AID area focused on feature engineering. In [21], the authors adopted a normalization technique as a preprocessor to improve the raw features. Improved incident detection results were obtained by applying the new features in a PNN classifier. Though this technique worked well, it is essentially a data normalization process under the AID context. We did not find feature learning study in the AID literature. Therefore, we believe there is a significant research gap in terms of feature learning in the AID context. In this research we propose to use unsupervised feature learning method to build higher level feature representation for incident detection data. By enhancing the raw features with higher level features, we expect to achieve better AID performance. Since learning feature mapping function is an unsupervised process, we also expect this new method shall potentially reduce the demand for labeled data in training better incident classifiers.

III. METHODOLOGY AND ALGORITHM

*A. Incident Data Used in the Study*

In this study we used real incident data collected in two sites namely I-405 northbound freeway and SR-22 eastbound freeway. Both of them are in the Orange County, California. The data contains real incidents in the entire year of 1998. Data includes both traffic volume and occupancy data averaged within the interval of 30 seconds. The data undertook the same pre-processing procedure as discussed in [5]. Generally speaking, the data contains multiple incident units. Each incident unit includes data for approximately 90 intervals (about 45 minutes) from both upstream and downstream detectors in which about 60 of them are data before the incident, the rest 30 of them are incident data and perhaps some ones after the incident. There is no guarantee that incident units in the data obey the chronological order. Since "non-capacity-reducing" incidents were removed from the original data as in [5], the resulting I-405 dataset has 52 incident units for training, 129 incident units for testing. In the training set, there are 4629 intervals in all, 1408 of them (30%) are incidents. In the testing set, there are 11445 intervals in all, 3699 of them (32%) are incidents. For the purpose of unsupervised feature learning, we did not use the labels of SR-22 dataset in this study. The SR-22 data we adopted in this study contains 4137 intervals in all.

Only four numbers are strictly corresponding to one detection moment in the data, namely volume and occupancy values for upstream and downstream respectively. Since such features are not sufficient in incident classification, a very common choice is to include four more intervals (t to t-4) from upstream detector and two more intervals (t to t-2) from downstream detector for both volume and occupancy (we call this choice [4-2] pair. We will generally abbreviate this as "[x-y] pair" later).

### B. Performance Measures

The following performance measures are used in this study. Most of them are conventional among many AID studies.

$$DR = \frac{Total\ number\ of\ detected\ incidents}{Total\ number\ of\ actual\ incidents} *100\% \quad (1)$$

$$FAR = \frac{Total\ number\ of\ false\ alarm\ cases}{Total\ number\ of\ intervals} *100\% \quad (2)$$

$$MTTD = \frac{Total\ time\ used\ to\ detect\ incidents}{Total\ number\ of\ detected\ incidents} *100\% \quad (3)$$

$$PI = (1.01 - DR) * (FAR + 0.001) * MTTD \quad (4)$$

Methods of computing detection rate (DR), FAR, MTTD and performance index (PI) are conventional as in many previous studies mentioned in the literature review. PI is a systematic measure which combines DR, FAR and MTTD [1][7]. This is used in the cross validation process as the optimization objective. We seek the lowest possible PI in the cross validation process. We slightly changed the equation of computing PI. Because DR could be 100% and FAR could be 0% during the training, it would cause invalid PI during cross validation if either case happens. We adopted the equation in (4) to avoid the possible exceptions caused by such cases. According to [22], any attempt to use a single number to represent multiple measures will lose some information. It is believed that PI is one of the best possible measures in the training process to reflect systematic AID performance in model selection.

### C. Unsupervised Feature Learning Algorithm

In this section, we will discuss the unsupervised feature learning algorithm and the ensuing incident classification process in more detail. The following six-step pipeline is used in this study to incorporate both unsupervised feature learning and incident classification. Similar pipeline was used in a number of unsupervised feature learning studies on various pattern recognition tasks [17].

1. Create an array with a number of $z+1$ dimensional unlabeled raw data vectors from the specified training data ($z=12$ in this study, we will explain why this is the case in part C of this section).

2. Extract random patches from unlabeled raw data vectors.

3. Use an unsupervised feature learning algorithm to learn a feature mapping function from the patches.

4. Prepare $z+1$ dimensional raw features from the training set.

5. Apply the learnt feature mapping function to sub-patches within each $z+1$ dimensional raw feature representation of every labeled detection moment. Then to generate the higher level features for labeled data using pooling [17].

   *(Repeat step 1 to step 5 when necessary, e.g. do it for volume and occupancy raw features respectively)*

6. Enhance the raw features with the newly generated higher level features and use the enhanced features to train a SVM classifier to classify incidents.

The input of step 1 is unlabeled data, it can either be the same dataset (left the labels unused) as which trains the incident classifier or data of other test sites. For example, if the unlabeled volume data from the upstream detector is $V = \{v^{(t-12)}, v^{(t-11)}, ..., v^{(t)}, v^{(t+1)}, ..., v^{(t+m)}\}$, where $v^{(i)}$ is a real number corresponding to the averaged volume at moment $i$, vector $\{v^{(t-12)}, ..., v^{(t)}\}$ is to be the first element of the array. In this case, there will be $m+1$ vectors in the entire array. In step 2, we randomly extract sub-patches from each vector. The dataset to be created after step 2 can be denoted as $X = \{x^{(1)}, x^{(2)}, ..., x^{(n)}\}$, where $x^{(i)} \in \Re^d$, $d \leq z$ and $d$ is the dimension of sub-patches. Both $d$ and $n$ are parameters to be specified in the experiment.

Though many unsupervised feature learning algorithms can be used in step 3, we purposefully used K-means algorithm in this study. K-means is not only an effective unsupervised feature learning algorithm [18], it is also straightforward to implement. Moreover, unlike other methods, K-means has only one parameter (number of centroids) to be fine-tuned in the feature learning stage. This fact significantly saves the training time. Since to our knowledge, this paper is the first one to study unsupervised feature learning in AID context, K-means algorithm is the ideal choice to work with.

In step 3, we apply K-means clustering to learn $K$ centroids $c^k$ from the input data, for example from $X$. Given the learnt centroids, we construct the following feature mapping function:

$$f_k(x`) = \max\{0, \mu(\tau) - \tau_k\} \quad (5)$$

where $\tau_k = \| x` - c^k \|_2$, $\mu(\tau)$ is the average of all the values in $\tau$, the input of the feature mapping function $x`$ is the raw feature for a detection moment to be prepared in step 4. The output of the feature mapping function is a $K$ dimensional higher level feature vector in which the $kth$ element is zero if the distance between $x`$ and $c^k$ is above average. This feature mapping function will generate higher level features with some degree of sparseness.

We then prepare the raw features from which the higher level features are extracted by using the feature mapping function (5) later on. Such raw feature is prepared for each detection moment and the dimension of them are $z+1$, same as unlabeled data vectors.

In step 5 we apply (5) for every sub-patch of $z+1$ dimensional raw feature vector from step 4 with the stride of one. For instance, if $z=2$ and sub-patch size is 2, then we will need to apply (5) twice to two possible sub-patches of the feature vector. The output of this step is a set of higher level features given by sub-patches of raw feature. We then use a very straightforward pooling strategy, simply adding them up, to generate the final higher level features.

It is noteworthy that step 1 through step 5 should be repeated four times since we need to do the same feature learning for volume and occupancy, both upstream and downstream.

The final step is to attach the higher level features to the raw feature and train a SVM classifier to carry out incident classification. We use a L1 regularized SVM with RBF kernel in this study. Model selection is done by 10-fold cross validation during the training (data is randomized by incident units in the cross validation to keep all the performance measures computable). The optimization objective during cross validation is PI.

### D. Further Data Pre-processing for This Study

In order to enable a completely fair comparison among all the experiments, we need to keep the number of incident and non-incident intervals the same across experiments for both training and testing set. However, by including more intervals in the raw features the number of total valid intervals will be decreased. For instance, if we include 6 intervals in the features, the 1st through 5th intervals of an incident unit will be invalid simply because there isn't enough data ahead. Therefore, certain amount of intervals at the top for each incident unit has to be excluded in the training and testing set to ensure such fair comparison. The number of data needs to be excluded for each incident unit equals $z$.

We specify $z=12$ in this study for three reasons. The first is we are able to find obvious performance pattern from [4-2] pair to [12-12] pair. The second reason is effective unsupervised feature learning could occur when the dimension of raw data is $z+1$. The third reason is after further data pre-processing using z equals 12, the total number of intervals in the training set is 4005 within which 1408 of them (35%) are incident data; in the testing set the number of total intervals are changed to 9897, within which 3699 of them (37%) represent incident data. Compare to the original data, the ratio between incident data and non-incident data is not significantly changed.

## IV. EXPERIMENTS AND ANALYSIS

### A. Finding Performance Patterns Using Raw Features

Despite most of the previous studies chose to use [4-2] pair, there is no prove to guarantee that other choices such as [4-3] pair or [8-8] pair would perform strictly better or worse. Therefore, we would like to know the performance dynamics when different choices of raw features are adopted in the incident classification. This information is important in this study mainly because the possible choices of raw features are potentially enormous and we would like to find particular performance patterns among these choices so that we can pick representative pairs for the later comparisons.

The first experiment was conducted by applying different choices of raw features in I-405 data, from [4-2] pair to [12-12] pair, and then directly feed into SVM for classification. We kept the number of intervals from upstream bigger or equals to the ones from downstream in the features. This aligns with the convention from previous studies. Same SVM as described in the previous section was used. PI was the optimization objective in cross validation during training. The following two figures illustrate the FAR and MTTD dynamics in the testing results.

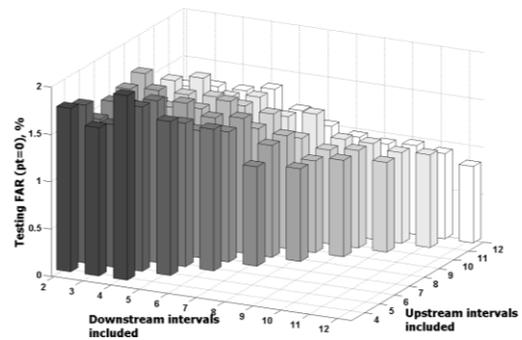

Figure 1.  *FAR dynamics using different raw feature choices*

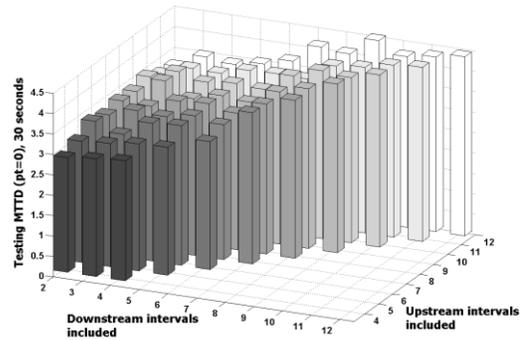

Figure 2.  *MTTD dynamics using different raw feature choices*

From figure 1, it's not hard to discern that FAR improves with more intervals included in the features. It is also obvious from figure 2 that MTTD performs worse with more intervals included in the features. There is no obvious pattern found in terms of DR. Therefore, we can conclude from this experiment that there might not be any choice of [x-y] pair performs strictly better than others. There is a trade-off between FAR and MTTD. We also found the interval choice along the diagonal is representative. Therefore, in the following experiment, we pick [4-2] pair, [8-8] pair and [12-12] pair for further comparisons with the enhanced features.

## B. Enhanced Features VS. Raw Features

We carried out the unsupervised feature learning process described in section 3 and compared the results with the three representative choices of raw features using I-405 dataset. When comparing with raw feature [x-y] pair, new features were constructed by combining [x-y] pair itself and the higher level features learnt by unsupervised feature learning to ensure the fairness[1]. We regard the enhanced feature better than the raw feature if it shall outperform the raw feature in all of the three representative cases. Due to the randomness of K-means algorithm, we repeated each experiment for 50 times. The final results were obtained by averaging the 50 results[2]. Figure 3 to figure 8 illustrate the experiment results. Values of DR and MTTD against different FAR values were obtained by using different number of persistent test (PT) [8].

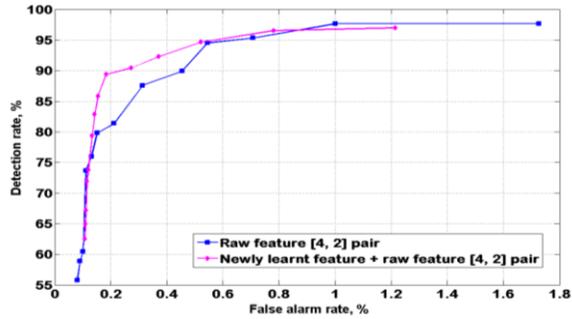

Figure 3.  *DR, raw feature [4-2] pair VS. enhanced features*

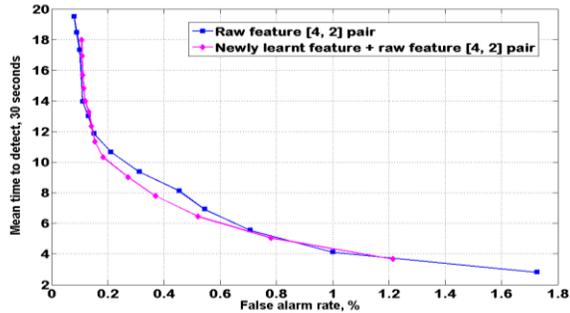

Figure 4.  *MTTD, raw feature [4-2] pair VS. enhanced features*

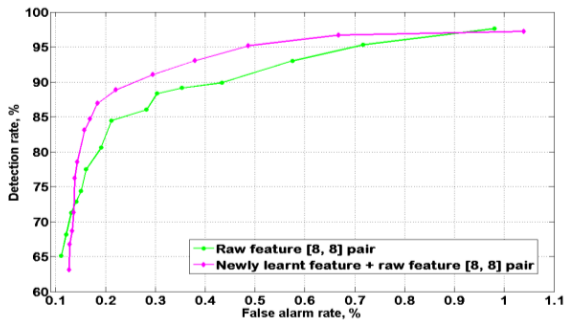

Figure 5.  *DR, raw feature [8-8] pair VS. enhanced features*

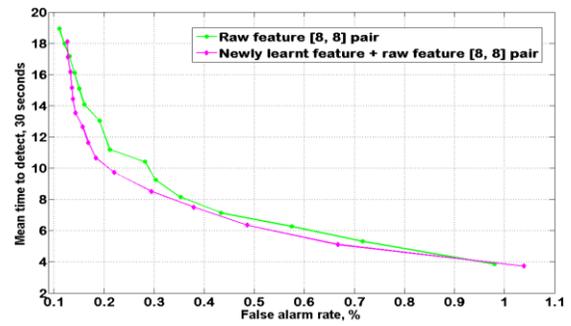

Figure 6.  *MTTD, raw feature [8-8] pair VS. enhanced features*

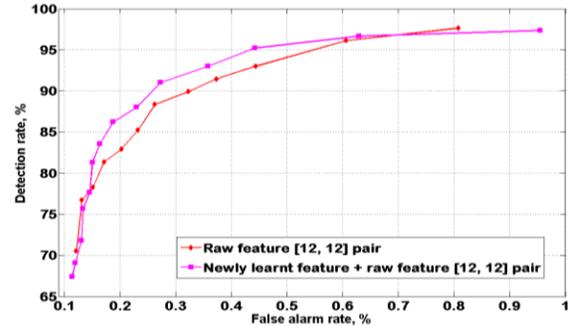

Figure 7.  *DR, raw feature [12-12] pair VS. enhanced features*

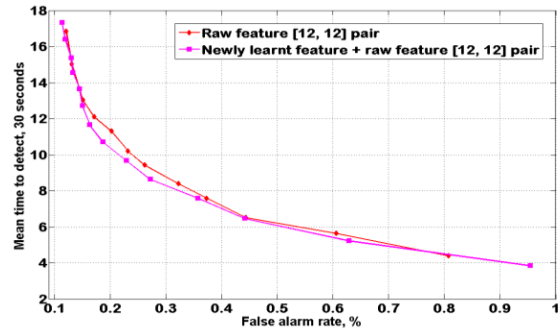

Figure 8.  *MTTD, raw feature [12-12] pair VS. enhanced features*

Several observations can be summarized from the results. First, by using the enhanced features, strictly better (higher) DR was obtained when FAR is ranged 0.1% to 0.7%. In some cases, for instance when FAR is 0.2% in figure 3, DR generated by the enhanced features is 10% higher than the one generated by raw features. Second, strictly better (lower) MTTD was also obtained for a large range of FAR by using the enhanced features. For example, we can observe from figure 6 that MTTD is improved by 45 seconds when FAR is around 0.25%. Third, by keeping the same DR and MTTD, better (lower) FAR is obtained.

## C. Learn Feature Mapping Function From Different Site

In the third experiment, we conducted the unsupervised feature learning described in section 3 by using the unlabeled data from SR-22 dataset to learn the feature mapping function (all other settings were the same as those in experiment 2). Then we extracted the higher level features from I-405 training set and trained the classifier. Testing was also performed using I-405 dataset. The results are listed in the following table in which case 1(b), case 2(b) and case 3(b) correspond to [4-2] pair based, [8-8] pair based and [12-12] pair based enhanced features. Case 1(a), case 2(a) and case 3(a) are their corresponding cases using the original I-405 dataset. Three rows in each performance measure show the results for pt=0 to pt=2.

---

[1] For instance, [4-2] pair itself was combined with higher level features to form the enhanced features when comparing with [4-2] pair raw feature.

[2] Number of centroids was 75 for volume, 15 for occupancy; patch size was 11 for volume, 6 for occupancy; number of patches sampled was 20000. Therefore, the dimension of the learnt higher level features was 180 (75*2+15*2). Due to the randomness of patch sampling and K-means, SVM parameter selected by cross validation was usually different among 50 times.

TABLE I.   RESULTS WHEN USING UNLABELED DATA FROM OTHER SITE

|  | Case 1(a) | Case 1(b) | Case 2(a) | Case 2(b) | Case 3(a) | Case 3(b) |
|---|---|---|---|---|---|---|
| DR (%) | 97.02 | 97.20 | 97.28 | 97.42 | 97.37 | 97.55 |
|  | 96.55 | 96.43 | 96.74 | 96.83 | 96.68 | 96.91 |
|  | 94.68 | 94.35 | 95.20 | 94.77 | 95.22 | 94.51 |
| FAR (%) | 1.215 | 1.274 | 1.039 | 1.100 | 0.960 | 0.975 |
|  | 0.782 | 0.812 | 0.668 | 0.683 | 0.631 | 0.628 |
|  | 0.521 | 0.536 | 0.487 | 0.504 | 0.445 | 0.454 |
| MTTD (30 seconds) | 3.669 | 3.709 | 3.731 | 3.777 | 3.944 | 3.877 |
|  | 5.066 | 5.031 | 5.117 | 5.122 | 5.218 | 5.296 |
|  | 6.465 | 6.441 | 6.352 | 6.369 | 6.447 | 6.458 |
| CR (%) | 87.14 | 87.03 | 88.19 | 88.11 | 88.25 | 88.17 |

CR stands for classification rate

The results are encouraging since we did not discern much difference if the data from SR-22 is used to learn the feature mapping function. If unlabeled data is not sufficient in one site, this method opens a door to enable feature learning from data elsewhere.

V. CONCLUSION AND FUTURE RESEARCH

We conclude the paper and summarize the contributions as the following. First, by using unsupervised feature learning we were able to construct enhanced features and improve AID performance for all the performance measures. Secondly, we showed that effective unsupervised feature learning for AID can not only take place by using the data from the same test site as that in incident classification, it is also valid to use unlabeled data from other test site. This obviously enriches the potential training data source. The direct consequence of these two results is that this approach may potentially reduce the number of labeled data required in training an incident classifier which achieves good detection performance. This is preferable because labeled AID data is expensive and difficult to get.

The limitations of this paper are two-fold. First, we only used one unsupervised feature learning method and didn't compare it with others. Second, we didn't visualize the higher level features and provide physical interpretations of them. This reduces the theoretical contribution of the study. We propose to work on these limitations in the future research.

To our knowledge, this is the first study to apply unsupervised feature learning in AID research. This approach shall not only provide an alternative way to enhance AID performance, we believe it shall also provide rich future research opportunities in the field. Other than the issues in the limitations, future research may also study how different parameters in the feature learning algorithm influence performance. Building deep architectures for AID is also within our study plan. Transferability of such algorithms is also an interesting future research direction.